\documentclass{article}

 \usepackage[preprint]{neurips_2026}

\usepackage[utf8]{inputenc} 
\usepackage[T1]{fontenc}    
\usepackage{hyperref}       
\usepackage{url}            
\usepackage{booktabs}       
\usepackage{amsfonts}       
\usepackage{nicefrac}       
\usepackage{microtype}      
\usepackage{xcolor}         
\usepackage[utf8]{inputenc} 
\usepackage[T1]{fontenc}    
\usepackage{hyperref}       
\usepackage{url}            
\usepackage{booktabs}       
\usepackage{amsfonts}       
\usepackage{nicefrac}       
\usepackage{microtype}      
\usepackage{xcolor}         
\usepackage{array}    
\usepackage{caption}
\usepackage{tabularx}
\usepackage{rotating}
\usepackage{tcolorbox}
\usepackage{algorithm}
\usepackage{algpseudocode}
\usepackage{amsmath}
\usepackage{algorithmicx}
\usepackage{booktabs}
\usepackage{multirow}
\usepackage{multicol}
\usepackage{amsfonts}
\usepackage{cleveref}
\usepackage{tabularx}
\usepackage{bbm}
\usepackage{wrapfig}
\usepackage{amssymb}
\usepackage{enumitem}
\setlist[itemize]{itemsep=0pt, topsep=0pt, parsep=0pt, partopsep=0pt}
\usepackage{graphicx}
\usepackage{subcaption}
\usepackage{lipsum} 

\title{Self-Consolidating Language Models:\\ Continual Knowledge Incorporation from Context}

\author{
    Zekun Wang\thanks{Equal contribution.} $\quad$ Anant Gupta\footnotemark[1] $\quad$ Zihan Dong $\quad$ Christopher J. MacLellan \\
    Georgia Institute of Technology\\
    \texttt{\{zekun,agupta886,zdong312,cmaclell\}@gatech.edu}
}

\begin{document}

\maketitle


\begin{abstract}
Large language models (LLMs) increasingly receive information as streams of passages, conversations, and long-context workflows. 
While longer context windows expose more evidence, they do not ensure that useful information is preserved and reused. 
We study continual context consolidation: writing current context into model weights while limiting interference with previously consolidated information.
We propose \textbf{S}elf-\textbf{Co}nsolidating \textbf{L}anguage Models (SCoL), a post-training framework in which, given current context, an LLM learns to generate textual update instructions specifying which of its own Transformer layers should be updated. 
Because committed updates change the model that later generates future selections, we train SCoL with meta-reinforcement learning over an evolving model state.
We instantiate SCoL with supervised QA rewards on SQuAD knowledge incorporation and intrinsic likelihood-based rewards for LongBench v2 long-context consolidation.
Across both settings, SCoL improves acquisition and retention over prompting, summarization, batch test-time training, and sequential finetuning baselines.
Analysis of learned selection patterns shows that SCoL encourages the LLM to generate sparse update locations that align with layers of high Fisher information, suggesting that the model learns to route plasticity toward loss-sensitive regions while limiting interference.
Moreover, SCoL transfers from shorter meta-training streams to longer LongBench v2 streams at evaluation, suggesting that our framework supports scalable streaming consolidation.
\end{abstract}
\section{Introduction}

Large language models increasingly operate over streaming context, including multi-turn conversations, tool-using agents, document workflows, and long-horizon interactive tasks.
In these settings, the model must not only answer from the current prompt, but also preserve earlier information for later use. 
Longer context windows help, but they do not guarantee reliable memory: models can underuse distant evidence and degrade as context length and reasoning complexity increase \cite{liu2023lost,hsieh2024ruler}.
Prior work manages long-context information through retrieval or external memory \cite{lewis2020rag,borgeaud2022retro,wang2023longmem}, attention-routing or memory mechanisms \cite{mohtashami2023landmark}, compression \cite{chevalier2023autocompressors,jiang2023llmlingua}, and agentic memory stores \cite{packer2023memgpt,zhong2023memorybank}. 
These methods are effective, but they often depend on retrieval quality, auxiliary modules, lossy compression, or repeated inference-time processing of persistent context.
We study a complementary direction: using the model weights as an internal memory substrate. Motivated by theories of memory consolidation, where experience is integrated into long-term internal representations rather than kept only as transient context \cite{mcclelland1995cls,kumaran2016cls,dudai2015consolidation}, we ask whether useful context can be written into parameters so it remains available after the original context is gone.

This reframes long context inference as a continual learning problem. The model must consolidate a stream of contexts into weights while preserving previously consolidated knowledge. However, sequential neural network updates are prone to catastrophic interference \cite{mccloskey1989catastrophic}. Classical methods reduce interference by estimating parameter importance, for example with Fisher based regularization \cite{kirkpatrick2017overcoming}, but reliable Fisher or Hessian style estimation is costly and implementation sensitive at LLM scale \cite{vandeven2025fisher}. We therefore ask whether the model itself can determine where adaptation should occur.

Inspired by SEAL \cite{zweiger2025seal}, we propose \emph{Self-Consolidating Language Models} (SCoL), a post-training framework in which a LLM learns to generate textual update instructions specifying with its own parameters to adapt for a given context.
We then attach LoRA adapters to the selected modules and perform a LoRA update \cite{hu2022lora}. 
Unlike SEAL, which focuses on generating self edits and update directives, our focus is where consolidation should occur under a drifting policy----each committed update changes the model that will choose future updates, so the policy itself is part of the evolving state. 
We trained SCoL with meta-reinforcement learning whose reward favors acquisition of the current context while penalizing forgetting of earlier contexts.


We instantiate SCoL under two reward settings: a supervised QA reward for SQuAD-style knowledge incorporation~\cite{rajpurkar2016squad}, and an intrinsic likelihood-based reward for unlabeled long-context streams evaluated on LongBench v2 \cite{bai2025longbenchv2}. Across both settings, SCoL improves acquisition and retention over in-context inference, summarization, batch test-time training, and sequential fine-tuning. Its learned selections are sparse and align with high-Fisher information layers, suggesting that SCoL routes plasticity toward loss-sensitive regions while limiting interference. On LongBench v2, selections meta-trained on shorter streams also transfer to longer evaluation streams without full-context access.

Our contributions are:
\begin{itemize}
\item We recast long-context inference as continual context consolidation, where useful information from a stream of contexts is written into model weights while preserving previously consolidated knowledge.
\item We proposed SCoL, a post-training framework that trains an LLM with meta-reinforcement learning to generate textual description of where in its own weights to update, and instantiate SCoL with supervised QA rewards and intrinsic likelihood-based rewards, covering both labeled knowledge incorporation and unlabeled long-context consolidation.
\item We evaluate against in-context inference, summarization, batch test-time training, and sequential fine-tuning baselines on SQuAD and LongBench v2, and show that teaching the LLM to generate update-location selections improves acquisition and retention, produces sparse selections that align with Fisher information, and transfers from shorter training streams to longer LongBench v2 context streams.
\end{itemize}
\section{Related work}
\label{sec:background}

\paragraph{Continual learning for large language models}
Continual learning studies how models acquire new tasks, domains, or knowledge without overwriting prior behavior. In LLMs, this appears in factual and lifelong model editing \cite{meng2022rome,mitchell2022mend,wang2024wise,chen2024lifelongediting}, continual domain adaptation \cite{gururangan2020dont}, and continual instruction or alignment tuning \cite{shi2024clsurvey,chen2026clllm}. Existing methods typically control forgetting by estimating parameter importance or preserving prior behavior through regularization \cite{kirkpatrick2017overcoming,li2016lwf}, or by restricting updates to low-rank, orthogonal, or modular parameter subspaces \cite{hu2022lora,wang2023olora,wang2024lemoe}.
These approaches are effective but often depend on fixed update rules, replay or distillation signals, hand-designed parameter partitions, or explicit importance estimates that can be costly at LLM scale \cite{vandeven2025fisher}. Our work instead learns a context-conditioned update policy: rather than computing weight importance directly, we post-trained the LLM to decide where new contextual knowledge should be inserted so adaptation improves incorporation while minimizing forgetting.

\paragraph{Inference with long context}
Long-context inference asks whether an LLM can reliably use information that appears far back in the prompt. Although context windows continue to grow, models can still underuse distant evidence and degrade as contexts become longer or more complex \cite{liu2023lost,hsieh2024ruler}. Existing approaches improve access to long-range information through retrieval or external memory \cite{lewis2020rag,guu2020realm,borgeaud2022retro,wang2023longmem}, special attention or memory mechanisms \cite{mohtashami2023landmark,wang2023longmem}, context compression \cite{rae2019compressive,chevalier2023autocompressors,jiang2023llmlingua}, or agentic memory systems \cite{zhong2023memorybank,packer2023memgpt,park2023generativeagents,shinn2023reflexion}.
These methods are effective but often keep information outside the model weights or compress it before reuse, making performance depend on retrieval quality, memory management, or lossy summarization. We study a complementary direction: consolidating context-derived knowledge into localized parameter updates. This reframes long-context inference as continual knowledge incorporation, where the model must retain useful information from a stream of contexts without overwriting previously consolidated knowledge.

\paragraph{Self-adaptation with reinforcement learning}
Reinforcement learning is often used to optimize LLMs as policies over text for alignment, reasoning, or task-level rewards \cite{ouyang2022instructgpt,guo2025deepseekr1}. In self-adapting LMs, the target instead shifts toward learning \emph{how the model should change} in response to new information, connecting to meta-learning and expert-iteration views of controlling an inner improvement process \cite{anthony2017exit}.
Recent work learns such adaptation strategies: Transformer-Squared learns task-conditioned self-adaptation with RL \cite{sun2025transformersquared}, and SEAL trains an LLM to generate self-edits and update directives \cite{zweiger2025seal}. CaMeLS is especially close to our setting because it meta-learns online adaptation over context, but it does so by training a separate small autoregressive model to assign token-level loss weights during fine-tuning \cite{hu2023camels}. 
By contrast, we proposed a post-training paradigm in which the LLM itself learns to propose \emph{where in its own weights} to update, without explicitly computing parameter importance. Since each committed update changes the model used for future contexts, we frame the problem as a meta-RL problem over a drifting model state, with the goal of continual knowledge consolidation under forgetting constraints.

\newcommand{\model}[1]{p_{#1}}
\newcommand{\thetab}{\theta}
\newcommand{\ctx}{c}
\newcommand{\cstream}{\mathcal{C}}
\newcommand{\cdset}{\mathcal{D}}
\newcommand{\qset}{\mathcal{Q}}
\newcommand{\pref}{\mathcal{P}}
\newcommand{\action}{a}
\newcommand{\mask}{m}
\newcommand{\refpolicy}{\pi_{\mathrm{ref}}}
\newcommand{\oplusop}{\,{\oplus}\,}
\newcommand{\Acq}{u}
\newcommand{\Forget}{f}
\section{Self-Consolidating Language Models}
\label{sec:method}

We study a continual consolidation setting in which a single LLM ingests a stream of contexts and must internalize each into its weights, without replay and without retrieval at inference time. 
\Cref{subsec:setup} formalizes continual context consolidation as a reinforcement learning problem whose reward decomposes into an acquisition term and a forgetting term.
\Cref{subsec:meta} presents SCoL, a meta-reinforcement learning procedure that trains the LLM to generate textual update-location selections under a drifting model state.
\Cref{subsec:sparse} instantiates the reward under supervised and sparse-supervision regimes.
The full procedure is given in \Cref{alg:seal-iter}.


\subsection{Problem setup: continual context consolidation}
\label{subsec:setup}

Let $\pi_{\thetab_0}$ denote a base language model (our policy) with parameters
$\thetab_0 \in \mathbb{R}^{d}$, and let
$\cstream = (\ctx_1, \ctx_2, \ldots, \ctx_T)$ be a stream of contexts, where each $\ctx_t \in \mathcal{X}^{*}$ is a token sequence.
We use \emph{context} generically: it can be a text passage, a multi-turn interaction history, or a long-context window.
At step $t$, the current model $\pi_{\thetab_{t-1}}$ receives $\ctx_t$ and generates a \textit{textual} action
\begin{equation}
    \action_t \sim \pi_{\thetab_{t-1}}(\cdot \mid \ctx_t),
    \label{eq:text-action}
\end{equation}
which is parsed into a structural update units, and in our main instantiation the units are Transformer layers.
We then update the current model by:
\begin{equation}
    \Delta \thetab_t
    =
    \mathrm{Adapt}(\thetab_{t-1}, \ctx_t, a_t),
    \qquad
    \thetab_t
    =
    \thetab_{t-1} \oplusop \Delta \thetab_t .
    \label{eq:state-transition}
\end{equation}
The prompting template and parsing procedure are described in \Cref{sec:app-cki}.
The weights $\thetab_t$ carry all accumulated history.
No retrieval index or auxiliary memory is maintained for generation.
During training, we keep a past-context set $\cdset_{<t}$ containing only the material needed to \textit{evaluate} forgetting from steps $1,\ldots,t-1$.
At inference time, the model relies on $\thetab_t$ alone.
After each consolidation step, the environment returns
\begin{equation}
    r_t(\thetab_t;\ctx_t,\cdset_{<t})
    =
    \Acq(\thetab_t;\ctx_t)
    -
    \lambda \Forget(\thetab_t;\cdset_{<t}),
    \label{eq:reward-abstract}
\end{equation}
where $\Acq$ measures acquisition of the current context, $\Forget$ measures drift on previously consolidated material, and $\lambda \geq 0$ trades off acquisition and retention.
We keep $\Acq$ and $\Forget$ abstract here and instantiate them in \Cref{subsec:sparse}.

The learning objective is to learn a \textit{pre-adaptation} policy $\pi_{\thetab_0}$ whose own sequential updates produce high cumulative reward:
\begin{equation}
    \max_{\thetab_0}\;
    \mathcal{J}(\thetab_0)
    =
    \mathbb{E}_{\substack{
        \action_t \sim \pi_{\thetab_{t-1}}(\cdot \mid \ctx_t)\\
        \thetab_t = \thetab_{t-1} \oplusop \mathrm{Adapt}(\thetab_{t-1},\ctx_t,\mask_t)
    }}
    \left[
        \sum_{t=1}^{T}
        r_t(\thetab_t;\ctx_t,\cdset_{<t})
    \right].
    \label{eq:cumulative-obj}
\end{equation}
The key point is that the policy parameters drift during the stream: the action at step $t$ is sampled from $\pi_{\thetab_{t-1}}$, whose parameters already contain all committed updates from earlier contexts.
Thus, the goal is not to learn a fixed context-to-update mapping, but to meta-learn an initialization whose rolled-out updates continue to produce good future update selections.

Several nearby paradigms share aspects of this problem but differ in a central architectural commitment.
RAG \citep{lewis2020rag} and memory-augmented architectures \citep{packer2023memgpt,chevalier2023autocompressors} handle new information by expanding an external store while leaving model parameters fixed.
Knowledge-editing methods \citep{meng2022rome,meng2023memit,mitchell2022mend,wang2024keditsurvey} target isolated factual updates rather than streaming accumulation and general context learning.
Batch fine-tuning \citep{howard2018ulmfit,raffel2020t5} requires simultaneous access to the full $\cstream$.
SCoL instead commits each incoming context into the model weights online, while learning where such updates should occur.

\subsection{Learning to learn without forgetting with meta-reinforcement learning}
\label{subsec:meta}

\begin{figure}
    \centering
    \includegraphics[width=1\linewidth]{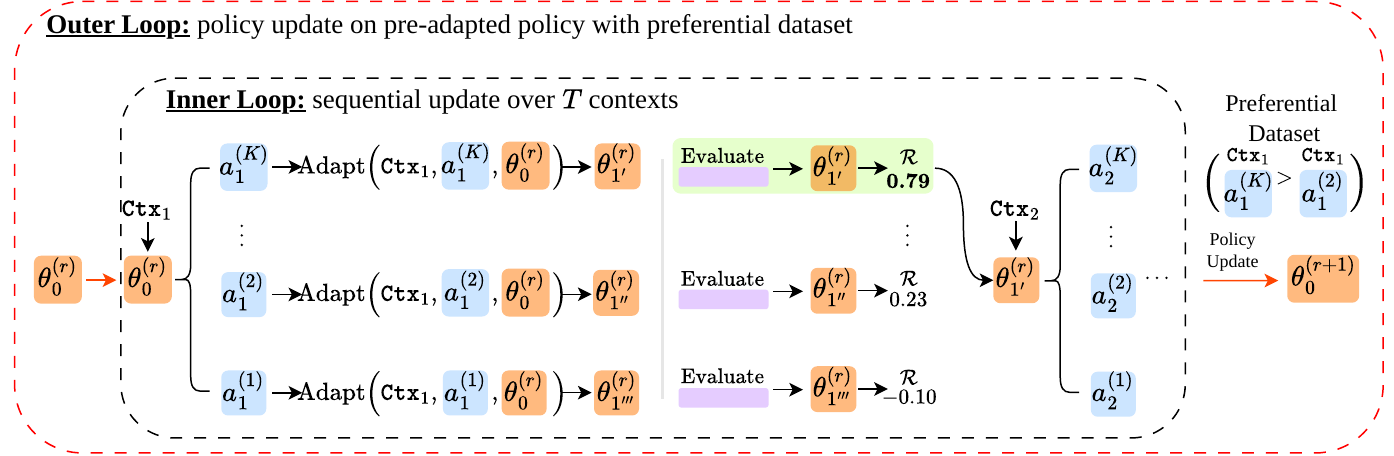}
    \caption{Training pipeline of the Self-Consolidating Language Model (SCoL).
    \textbf{Inner loop (black):} at each context $\mathrm{Ctx}_t$, the current model state $\theta^{(r)}_{t-1}$ samples $K$ textual actions, adapts a candidate model, and scores it with the reward in \Cref{eq:reward-abstract}. The highest-reward candidate is committed, advancing the running model state.
    \textbf{Outer loop (red):} after the stream is exhausted, the induced preferential dataset trains the round-start policy $\theta_0^{(r)}$ with IPO, producing $\theta_0^{(r+1)}$ for the next round.}
    \label{fig:itc_pipeline}
\end{figure}


SCoL parameterizes the adaptation decision as text generation.
For each context, the LLM emits a textual list of update locations. We then attach LoRA modules \citep{hu2022lora} to the selected structural units and perform a CLM update on $\ctx_t$:
\begin{equation}
    \Delta\thetab^{\star}(a_t)
    =
    \arg\min_{\Delta\thetab_{a_t}}
    \mathbb{E}_{x \sim \ctx_t}
    \bigl[
        -\log \model{\thetab_{t-1}+\Delta\thetab_{a_t}}(x)
    \bigr].
    \label{eq:inner-adapt}
\end{equation}
\begin{wrapfigure}[22]{r}{0.48\textwidth}
\vspace{-14px}
\begin{minipage}{0.48\textwidth}
\captionof{algorithm}{\textbf{Self-Consolidating LM Training:}\\ Learning to Learn Without Forgetting}
\label{alg:seal-iter}
\hrule\vspace{2pt}
{\footnotesize
\linespread{0.9}\selectfont
\begin{algorithmic}[1]
\Require $\theta_0^{(1)}$, stream $\mathcal{C}=(c_1,\ldots,c_T)$, $K$, $\lambda$, rounds $R$
\Ensure trained policy $\pi_{\theta_0^{(R+1)}}$
\For{$r = 1,\ldots,R$}
    \State $\theta^{(r)}_0 \gets \theta^{(r)}_0$;\quad $\mathcal{P}^{(r)} \gets \varnothing$
    \For{$t = 1,\ldots,T$}
        \State  $a_t^{(1)},\ldots,a_t^{(K)} \sim \pi_{\theta^{(r)}_{t-1}}(\cdot\mid c_t)$
        \For{$k = 1,\ldots,K$}
            \State $\Delta_t^{(k)} \gets \textsc{Adapt}(\theta^{(r)}_{t-1}, c_t, a_t^{(k)})$
            \State $\theta_t^{(r,k)} \gets \theta^{(r)}_{t-1} \oplusop \Delta_t^{(k)}$
            \State $r_t^{(k)} \gets u(\theta_t^{(r,k)}; c_t)-\lambda f(\tilde{\theta}_t^{(r,k)};\mathcal{D}_{<t})$
        \EndFor
        \State append $\{(a^w,a^l,c_t):r_w>r_l\}$ to $\mathcal{P}$
        \State $k^\star \gets \arg\max_k r_t^{(k)}$
        \State $\theta^{(r)}_t \gets \tilde{\theta}_t^{(r,k^\star)}$
    \EndFor
    \State $\theta_0^{(r+1)} \gets \arg\min_{\theta}\mathcal{L}_{\mathrm{IPO}}(\theta;\mathcal{P}^{(r)})$
\EndFor
\State \Return $\pi_{\theta_0^{(R+1)}}$
\end{algorithmic}
\par}
\vspace{2pt}\hrule
\end{minipage}
\vspace{-\baselineskip}
\end{wrapfigure}
This yields the candidate post consolidation model
\begin{equation}
    \thetab_t^{(\action)}
    =
    \thetab_{t-1} \oplusop \Delta\thetab^{\star}(a).
    \label{eq:candidate-state}
\end{equation}
Across candidates, the inner-loop hyperparameters are fixed; candidates differ only in which structural units receive adapters.
Thus, the learned decision is where to adapt, not how strongly to adapt.
At each context, we sample $K$ candidate textual actions from the current, already-drifted policy:
\begin{equation}
    \action_t^{(1)},\ldots,\action_t^{(K)}
    \sim
    \pi_{\thetab_{t-1}}(\cdot \mid \ctx_t).
    \label{eq:k-samples}
\end{equation}
Each action is parsed, adapted, and scored:
\begin{equation}
    r_t^{(k)}
    =
    r_t\!\left(
        \thetab_t^{(\action_t^{(k)})};
        \ctx_t,\cdset_{<t}
    \right).
    \label{eq:candidate-reward}
\end{equation}
We then commit the highest-reward candidate,
\begin{equation}
    k^{\star}
    =
    \arg\max_{k} r_t^{(k)},
    \qquad
    \thetab_t
    \gets
    \thetab_t^{(\action_t^{(k^{\star})})}.
    \label{eq:commit-best}
\end{equation}
This committed update changes the model that will generate the next action: $\pi_{\thetab_t}$ is both the consolidated model and the policy used at the next context.
The $K$ candidate scores also induce preference data.
For each context, we construct
\begin{equation}
    \pref_t
    =
    \left\{
        (\ctx_t,\action^w,\action^l)
        :
        r_t(\action^w) > r_t(\action^l)
    \right\},
    \label{eq:pref-pairs}
\end{equation}
where $\action^w$ and $\action^l$ denote higher- and lower-reward textual actions for the same context.
Across the full inner stream, we collect $\pref^{(r)}=\bigcup_{t=1}^{T}\pref_t$.

After a full round of $T$ sequential consolidations, we update the \emph{pre-adaptation} parameters $\thetab_0^{(r)}$, not the final drifted parameters $\thetab_T^{(r)}$.
This is the meta-learning step: the preference data are generated by rolling the current policy through a stream of self-induced updates, but the outer update improves the round-start model so that its future rollouts generate better update-location selections.
We use Identity Preference Optimisation (IPO) \citep{azar2024ipo}:
\begin{equation}
    \mathcal{L}_{\mathrm{IPO}}(\theta)
    =
    \mathbb{E}_{(\ctx,\action^w,\action^l)\sim \pref^{(r)}}
    \left[
        \left(
            \log\frac{\pi_\theta(\action^w \mid \ctx)}
                     {\pi_{\mathrm{ref}}(\action^w \mid \ctx)}
            -
            \log\frac{\pi_\theta(\action^l \mid \ctx)}
                     {\pi_{\mathrm{ref}}(\action^l \mid \ctx)}
            -
            \frac{1}{2\beta}
        \right)^2
    \right],
    \label{eq:ipo}
\end{equation}
with $\pi_{\mathrm{ref}}=\pi_{\thetab_0^{(r)}}$ fixed to the round-start policy.
The minimizer becomes the next round-start policy,
\begin{equation}
    \thetab_0^{(r+1)}
    =
    \arg\min_{\theta}
    \mathcal{L}_{\mathrm{IPO}}(\theta).
    \label{eq:outer-update}
\end{equation}
The next round begins from $\thetab_0^{(r+1)}$ and produces a fresh drifting trajectory $\thetab_1^{(r+1)},\ldots,\thetab_T^{(r+1)}$.
We use IPO rather than supervised fine-tuning on the top-1 action because the action space consists of low-entropy structured strings.
In preliminary experiments, SFT-style reinforcement such as ReST \citep{gulcehre2023rest} quickly concentrated probability mass on a few repeated layer lists.
IPO instead stops updating once the chosen--rejected log-ratio reaches the finite margin $1/(2\beta)$, which helps preserve diversity among plausible update-location selections.
Additional comparisons between outer-loop update algorithms are provided in \Cref{app:outer_algo_sweep}.

\subsection{Reward instantiations and reward sparsity}
\label{subsec:sparse}
The abstract reward of \Cref{eq:reward-abstract} admits two instantiations, distinguished by whether a context arrives with downstream supervision.

When $\ctx_t$ is accompanied by a set of held-out queries
$\qset_t = \{(q, a)\}$ tied to the context, both reward terms are measured
in the same units of downstream performance. 
The acquisition term is the accuracy on $\qset_t$ of the candidate post-consolidation model. 
The forgetting term is the accumulated degradation on past query sets, measured against each past context's own first-consolidation baseline.
Concretely, for each past step $s < t$ we cache the accuracy
$b_s = \mathrm{Acc}_{\qset_s}(\thetab_{s+1})$ attained immediately after
$\ctx_s$ was first consolidated, and we penalise any drop from this
baseline under the candidate state $\thetab_t^{(\action)}$. That is,
$\cdset_{<t} = \{(\ctx_s, \qset_s, b_s)\}_{s<t}$ in this regime, and
\begin{equation}
    r\bigl(\action; \ctx_t, \thetab_t\bigr)
    \;=\;
    \mathrm{Acc}_{\qset_t}\!\bigl(\thetab_t^{(\action)}\bigr)
    \;-\;
    \frac{\lambda}{t-1}
    \sum_{s=1}^{t-1}\!
    \Bigl[
        b_s \;-\; \mathrm{Acc}_{\qset_s}\!\bigl(\thetab_t^{(\action)}\bigr)
    \Bigr].
    \label{eq:reward-labeled}
\end{equation}

Many settings of interest do not provide such supervision. Contexts stream in without labels; downstream performance is unavailable, noisy, or delayed. 
In this regime $\cdset_{<t} = \{\ctx_s\}_{s<t}$, and we instantiate both reward terms intrinsically. 
A well-consolidated model $\thetab^{\prime}$ should raise the likelihood of $\ctx_t$ relative to the pre-adaptation model $\thetab_t$,
\begin{equation}
    \Acq_{\mathrm{intrinsic}}\!\left(\thetab^{\prime}; \ctx_t\right)
    \;=\;
     \log \model{\thetab^{\prime}}(\ctx_t)
    \;-\;
    \log \model{\thetab_t}(\ctx_t),
    \label{eq:acq-intr}
\end{equation}
and maintain the likelihood of $x \in \mathcal{D}_{<t}$ as measured by the output-distribution drift on previously-consolidated material. 
Because $\cdset_{<t}$ is an empirical collection of past contexts rather than samples drawn from the model, absolute log-likelihood differences can disproportionately weight passages with higher baseline likelihood or lower entropy, leading to uneven retention where some contexts dominate the forgetting signal. To ensure uniform retention across previously consolidated material, we instead measure forgetting in relative terms by normalizing the change in log-likelihood by its pre-adaptation value. 
This yields a notion of fractional degradation that treats each passage comparably, regardless of scale, and better aligns with the objective of preserving all contexts without bias. 
Concretely, we define
\begin{equation}
    \widetilde{\Forget}_{\mathrm{intrinsic}}\!\left(\thetab^{\prime}; \cdset_{<t}\right)
    \;=\;
    \mathbb{E}_{x \sim \cdset_{<t}}\left[
        \frac{
            \log \model{\thetab_t}(x) - \log \model{\thetab^{\prime}}(x)
        }{
            \bigl|\log \model{\thetab_t}(x)\bigr|
        }
    \right] = \mathbb{E}_{x \sim \cdset_{<t}}\left[\frac{\Acq_{\mathrm{intrinsic}}\!\left(\thetab^{\prime}; x\right)}{\log \model{\thetab_t}(x)}\right],
\end{equation}
While this normalization ensures that each passage contributes comparably, it removes the original scale of log-likelihood differences. To restore a meaningful magnitude to the forgetting signal, we rescale by the average pre-adaptation log-likelihood over $\cdset_{<t}$, yielding
\begin{equation}
    \Forget_{\mathrm{intrinsic}}\!\left(\thetab^{\prime}; \cdset_{<t}\right)
    \;=\;
    \widetilde{\Forget}_{\mathrm{intrinsic}}\!\left(\thetab^{\prime}; \cdset_{<t}\right)
    \cdot
    \mathbb{E}_{x \sim \cdset_{<t}}\left[
        \bigl|\log \model{\thetab_t}(x)\bigr|
    \right],
    \label{eq:forget-intr}
\end{equation}
In both regimes the forgetting term is a drop relative to a per-context baseline evaluated on the same test material, differing only in whether that material is labeled queries or the raw context. 
Substituting \Cref{eq:acq-intr,eq:forget-intr} into \Cref{eq:reward-abstract} yields
\begin{equation}
    r_{\mathrm{sparse}}\!\bigl(\action; \ctx_t, \thetab_t\bigr)
    \;=\;
    \Acq_{\mathrm{intrinsic}}\!\bigl(\thetab_t^{(\action)}; \ctx_t\bigr)
    \;-\;
    \lambda\,\Forget_{\mathrm{intrinsic}}\!\bigl(\thetab_t^{(\action)}; \cdset_{<t}\bigr),
    \label{eq:reward-sparse}
\end{equation}
applicable to rolling-window long-context consolidation
\citep{chen2023positional,chevalier2023autocompressors}, memory-compaction
in agentic interaction histories \citep{packer2023memgpt,mu2023gisting},
and any setting in which a context arrives unlabeled. 

We then evaluate our self-consolidating LMs framework under the above two reward instantiations: context that comes with immediate downstream supervisions (\Cref{sec:exp-cki}), and context that has no downstream supervisions (\Cref{sec:exp-lcs}).


\section{Continual knowledge injection}
\label{sec:exp-cki}

\subsection{Experiment setup}
\paragraph{Task and dataset.}

We evaluate continual knowledge incorporation, in which a language model is updated in sequence on a stream of SQuAD \citep{rajpurkar2016squad} passages and held responsible for answering a short question-answer set attached to each passage. 
We prepare a disjoint training split used for meta-training the policy and a validation split reserved for sequential-update evaluation. \begin{wrapfigure}[18]{r}{0.35\textwidth}
    \centering
    \vspace{-12px}
    \includegraphics[width=0.35\textwidth]{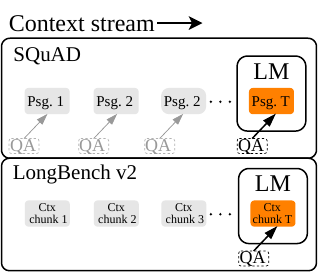}
    \caption{Illustrations of experiment setup under two reward settings. \textbf{Top}: each context (SQuAD passages) comes with an environment reward (downstream QA). \textbf{Bottom}: QA is evaluated at the end of the context stream.}
    \label{fig:reward}
    \vspace{-1em}
\end{wrapfigure}
\hspace{-0.1cm}Additional dataset details are provided in \Cref{sec:app-stream}.

\paragraph{Policy, inner adaptation, and meta update.}
We choose the policy as a \texttt{Qwen2.5-7B-Instruct} model: at each context, the model is prompted to emit a list of up to $B = 10$ transformer-layer indices. 
Additional results for finer granularities (i.e, projection layers) are discussed in \Cref{app:granularity_comparison}.
Our inner adaptation follows SEAL setting: for every candidate action, we attach a LoRA adapter to the selected layers and run the CLM update of \Cref{eq:inner-adapt} on a set of implications pre-generated once per passage by a frozen SEAL-pretrained model, for 10 epochs at learning rate $2 \times 10^{-4}$ and batch size $1$. 
During meta-training we stream $T = 50$ passages per round; at each passage we sample $K = 10$ candidates actions from the current policy, score each candidate under of \Cref{eq:reward-labeled} using a frozen \texttt{Qwen2.5-7B-Instruct} as a judge, commit the argmax to the running weights per \Cref{eq:commit-best}, and accumulate pairwise comparisons of reward scores into a round-level buffer.
At the end of each round the buffer is used to reinforce the \textit{pre-adaptation} base policy with IPO (\Cref{eq:ipo}), producing the base policy that seeds the next round; we report $R = 2$ such rounds. 
During evaluation, LLM only emits the selection once (i.e. \(K=1\)) and commits the update.
Prompts and additional training detailes are in \Cref{sec:app-selection-prompt,sec:app-impl-prompt,sec:app-inner-lora,sec:app-outer-ipo,sec:app-reward-judge}.

\paragraph{Baselines and metrics.}

We compare against three baselines and three SCoL variants. All methods use the same pre-generated implications and inner LoRA update procedure, so the comparison isolates the effect of \emph{where} adaptation is applied.
\textbf{QA Prompting} prompts the base model with the question only. 
\textbf{Batch TTT} trains a LoRA adapter jointly on all evaluation passages in an offline pass, serving as a batch-access upper bound. 
\textbf{SEAL}$_{\text{continual}}$ applies the SEAL knowledge-incorporation procedure to the continual stream and serves as the continual LoRA fine-tuning baseline. 
$\textbf{SCoL}_{\lambda=0}$ removes the forgetting term from our reward, isolating its contribution. 
\textbf{SCoL} uses the full reward with $\lambda=1$. 

At evaluation we apply each method to a fixed validation stream of $N = 100$ passages and record the sequential-update accuracy matrix $M \in [0,1]^{N \times N}$ whose entry $M_{i,j}$ is the accuracy on the $j$-th passage's queries after the model has been edited through the $i$-th passage (for $j \le i$). 
From $M$ we report two metrics for each round: \textbf{immediate acquisition}, the mean of the diagonal, measuring how well each update works at the moment it is committed and \textbf{retention}, the mean of the final-row entries on past passages, measuring accuracy on everything learned earlier once the full stream has been processed. 
Metric formulas and baseline details are given in \Cref{sec:app-metrics,sec:app-baselines}.

\subsection{Results and discussions}
\begin{figure}[t]
    \centering

    \begin{minipage}[t]{0.24\textwidth}
        \vspace{0pt}
        \centering
        \includegraphics[width=\linewidth]{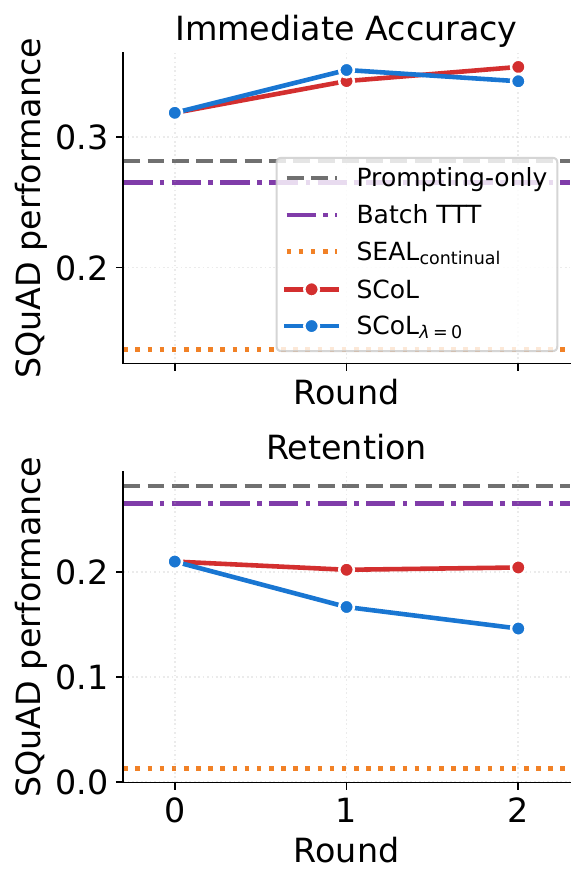}
        \label{fig:performance_layer}
    \end{minipage}
    \hfill
    \begin{minipage}[t]{0.75\textwidth}
        \vspace{0pt}
        \centering

        \includegraphics[width=\linewidth,trim=0 0 0 0,clip]{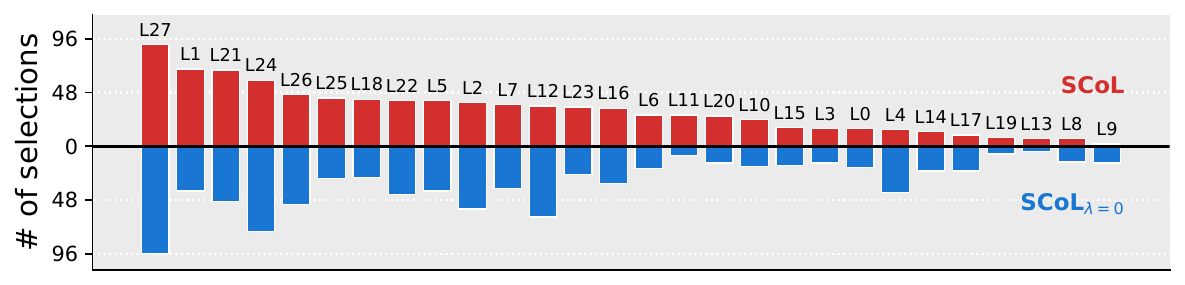}
        \label{fig:selection_count_layer}

        \vspace{-1.5em}

        \includegraphics[width=\linewidth,trim=0 0 0 0,clip]{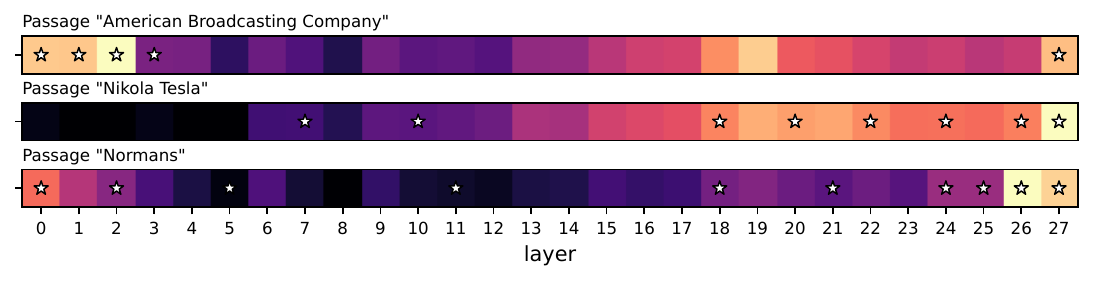}
        \label{fig:fisher-alignment}
    \end{minipage}

    \vspace{-0.5em}
    \caption{\textbf{Left:} immediate accuracy (top) and retention (bottom) across IPO updates. \textbf{Top Right:} top layer selection frequencies after the final IPO update, with the full reward variant in red and the acquisition only variant in blue. \textbf{Bottom Right:} examples of per-layer weight importance measured by Fisher information (brighter is higher) with SCoL's selected layers marked by white stars.}

    \label{fig:performance_and_layer_analysis}
    \vspace{-0.5em}
\end{figure}

Table~\ref{tab:method_retention} shows that both SCoL and SCoL$_{\lambda=0}$ exceed all three baselines in immediate accuracy, indicating that the learned policy improves acquisition beyond prompting, batch TTT, and sequential updating. 
Batch TTT performs worse than prompting only, despite seeing all 100 passages and implications jointly, with an immediate accuracy drop from 28.17\% to 26.52\%.
We interpret this as a declarative training, query evaluation mismatch. 
The adapter is trained on passage text and implication strings under language modeling, but is later queried through QA. 
This resembles the reversal curse \citep{berglund2024reversal}, where facts learned in one direction are not reliably retrieved in another.
SEAL collapses under sequential updating, consistent with compounding drift in parameter efficient continual tuning \citep{huang2024analyzing,wang2023olora}.
Retention clarifies the role of the forgetting term. 
Prompting only and Batch TTT are no update reference points whose retention equals to their immediate score. \begin{wraptable}[13]{r}{0.48\textwidth}
\centering
\vspace{-0.5em}
\small
\setlength{\tabcolsep}{3pt}
\renewcommand{\arraystretch}{1.2}
\caption{Comparison of immediate accuracy and retention across methods on continual knowledge incorporation over 100 SQuAD passages.}
\begin{tabular}{lcc}
\toprule
\textbf{Method} & \textbf{Immediate Acc.} & \textbf{Retention} \\
\midrule
Prompting-only & 28.17\% &--\\ 
Batch TTT & 26.52\% & --\\
\text{SEAL}$_{\text{continual}}$ & 13.70\% & 1.30\% \\
\hline
$\textbf{SCoL}_{\lambda=0}$ & 34.30\% & 14.64\% \\
$\textbf{SCoL}$ & \textbf{35.40\%} & \textbf{20.46\%} \\
\bottomrule
\end{tabular}

\label{tab:method_retention}
\end{wraptable}
\hspace{-0.1cm}SEAL retains only 1.30\% accuracy after sequential incorporation. 
In contrast, SCoL$_{\lambda=0}$ achieves stronger immediate acquisition, but its retention remains limited at 14.64\%. 
Using forgetting reward $f$ recovers retention to 20.46\% while also improving immediate accuracy. 
This suggests that explicitly rewarding retention can substantially reduce catastrophic forgetting in a self-improving language model framework \cite{zweiger2025seal}.
Figure~\ref{fig:performance_and_layer_analysis} further shows that retention decays monotonically without $f$, whereas SCoL keeps retention nearly stationary and continues improving acquisition. 
The layer selection patterns in Figure~\ref{fig:performance_and_layer_analysis} indicate that the learned policy is not selecting arbitrary update sites. 
Both SCoL and SCoL$_{\lambda=0}$ concentrate their selections in the last layer (\(L27\)).
This agrees with prior evidence that late blocks in autoregressive Transformers act as storage and retrieval sites for factual associations \citep{geva2021kvmem,meng2022rome}. 
We further test a fixed-selection ablation that always updates the last 10 layers. 
It achieves \(28.78\%\) immediate accuracy and \(20.10\%\) retention, close to but below SCoL's \(20.46\%\). 
This supports the view that late layers preserve factual associations, and that SCoL learns to exploit this structure, while its stronger immediate accuracy suggests that context-dependent layer selection is more robust than a fixed late-layer strategy.
At the same time, both variants avoid several early and middle layers, with \(L8\) and \(L9\) rarely selected and changing little across IPO rounds (see \Cref{fig:selection_delta}).
We refer to Appendix~\ref{app:addi-results-cki} for additional results and standard errors.

\paragraph{Fisher alignment.}
We then ask what the LLM is learning to select, and how these generated layer selections contribute to retention. 
Since Fisher information measures where the current passage loss is most sensitive in the running model, it provides a direct diagnostic for whether SCoL routes adaptation toward steep layerwise directions while keeping updates sparse. 
We measure layerwise Fisher information on the running model's base projection weights from each passage's implication lines, with details in Appendix~\ref{app:fisher-computation}.
Given the layer selection set \(S(c)\) generated for context \(c\), we measure alignment as \(\mathrm{recall@}k(c)=|S(c)\cap \mathrm{Top}_{k(c)}(F(c))|/k(c)\), where \(k(c)=|S(c)|\) and \(F(c)\) is the layerwise Fisher vector.
A random selection of 10 layers (out of 28 total layers) has expected recall \(10/28=35.7\%\). 
\begin{wraptable}{r}{0.45\textwidth}
\centering
\caption{Layerwise Fisher alignment on 100 SQuAD passages. Values report \texttt{recall@k(c)} against the Fisher top-$k(c)$ layers for learned and random selections.}
\label{tab:fisher-alignment}
\vspace{0.3em}
\small
\setlength{\tabcolsep}{2.5pt}
\renewcommand{\arraystretch}{1.2}
\begin{tabular}{lcc}
\toprule
       & \multicolumn{2}{c}{\texttt{recall@k(c)}} \\
\cmidrule(lr){2-3}
Method & Learned selection & Random selection \\
\midrule
SCoL & $\mathbf{45.1}_{\pm 1.6}$ & $35.7$ \\
SCoL$_{\lambda=0}$ & $41.0_{\pm 1.3}$ & $35.7$ \\
\bottomrule
\end{tabular}
\vspace{-1.0em}
\end{wraptable}
\hspace{-0.1cm}Table~\ref{tab:fisher-alignment} shows that SCoL reaches \(45.1\%_{\pm 1.6\%}\), while SCoL\(_{\lambda=0}\) reaches \(41.0\%_{\pm 1.3\%}\), both clearly above the random selection baseline. 
This provides evidence that the learned selections are not merely produced by the prompt format and do not degenerate to random sparse choices, and shows that IPO reinforces the LLM to generate update locations that follow high sensitivity layers for the current passage. 
Figure~\ref{fig:performance_and_layer_analysis} shows SCoL selections overlaid on Fisher information heatmaps for sampled SQuAD passages, where the generated layer selections concentrate on high (bright) Fisher layers.
As a result, SCoL updates layers that are most responsive to the current passage, but does so under a sparse selection pattern that leaves most layers untouched. 
Such selective plasticity matches the principle behind continual learning methods that reduce catastrophic interference by restricting adaptation to sparse, task specific update pathways \cite{zenke2017continual,mallya2018packnet,serra2018overcoming}.

\section{Long-context consolidation}
\label{sec:exp-lcs}

\subsection{Experiment setup}
\paragraph{Task, dataset, and evaluation.}
In many real world settings, models often receive information as a continuous stream (e.g., daily updates to knowledge base) without intermediate supervision, and must answer queries based on accumulated context.
To simulate this setting, we use LongBench-v2~\citep{bai2025longbenchv2}, where each example consists of a long context paired with a question requiring reasoning over the full passage.
We treat each passage as a sequential stream of fixed-length (2048-token) segments. The objective is to consolidate information from the stream into model parameters, such that the model can answer the final query without access to the original context.
To evaluate both consolidation and generalization, we consider two regimes: (1) \textbf{short}, with context lengths of 16k--32k tokens, and (2) \textbf{long}, with context lengths of 32k--64k tokens. The model is trained only on the short regime and evaluated on both.
We evaluate on $40$ short passages and $20$ long passages. Due to the sparsity of the reward signal, evaluation is performed by sampling 10 candidate answers after consolidation and measuring mean accuracy. Additional details are provided in \Cref{app:lcs-imp-det}.
\paragraph{Training procedure.}
We use \texttt{Qwen2.5-7B-Instruct} as our policy and reuse hyperparameters from \Cref{sec:exp-cki}. Meta-training consists of 2 rounds, each processing 5 passages (or streams). For each passage, at each step, we sample $K=10$ candidate actions, perform inner-loop adaptation and use the intrinsic reward defined in \Cref{eq:reward-sparse} to select the highest-scoring candidate. The selected adapter is merged into the running model (\Cref{eq:commit-best}) before proceeding to the next segment.

\paragraph{Baselines.}
We compare methods across three settings: \textbf{(1) In-context.} \textbf{Base}: the base model answers questions without additional context. \textbf{Full Context}: the base model is provided the entire passage. \textbf{Summarization}: the base model is given a concise summarization of the passage by \texttt{Kimi-K2.6}~\citep{kimi_k2_6}. \textbf{(2) Test-time training.} \textbf{Batch TTT}: a single LoRA adapter is trained on the full passage before querying. \textbf{Sequential FT}: Sequentially LoRA Fine-Tuning the model over passage segments without any module selection. \textbf{(3) Self-consolidation.} \textbf{SCoL}: our framework.

\subsection{Results and discussions}
\begin{wraptable}[14]{r}{0.35\textwidth}
\centering
\vspace{-15px}
\small
\setlength{\tabcolsep}{5pt}
\renewcommand{\arraystretch}{1.2}
\caption{Long-context consolidation comparison across two context-length regimes (short: 16k--32k and long: 32k--64k tokens).}
\begin{tabular}{lcc}
\toprule
\textbf{Method} & \textbf{short} & \textbf{long} \\
\hline
\noalign{\vskip 1pt}
Base        & 32.5\% & 29.5\% \\
Full Context    & 38.0\% & 35.5\% \\
Summarization   & 39.3\% & 36.0\% \\
\hline
\noalign{\vskip 1pt}
Batch TTT           & \underline{39.5\%} & \textbf{41.5\%} \\
Sequential FT       & 36.5\% & 23.0\% \\
\hline
\noalign{\vskip 1pt}
\textbf{SCoL}     & \textbf{42.3\%} & \underline{37.0\%} \\
\bottomrule
\end{tabular}
\label{tab:lcs-results}
\end{wraptable}

 
\Cref{tab:lcs-results} summarizes performance across both context regimes. SCoL achieves the strongest performance in the short-context setting at 42.3\%, outperforming all baselines. In the long-context regime, SCoL remains competitive at 37.0\%, significantly surpassing Sequential FT and exceeding in-context methods. These results demonstrate that explicit consolidation into model weights effectively integrates information across context lengths without requiring full access to the sequence.
Batch TTT serves as a performance upper bound, achieving 41.5\% in the long-context regime due to its global optimization over the entire sequence.
In contrast, Sequential FT exhibits a performance collapse as context length increases ($36.5\% \rightarrow 23.0\%$), underscoring the challenge of catastrophic forgetting under naïve online updates. SCoL consistently outperforms Sequential FT, confirming that structured consolidation is critical for stable continual adaptation.
In the in-context setting, providing full context or a Kimi-K2.6 summary improves over the base model, but these gains diminish as context length increases. This degradation, consistent with prior findings that context length alone can hurt performance \citep{du2025context}, emphasizes the limitations of relying solely on context or lossy summarization and motivates the transition to persistent parameter-based consolidation.

Overall, weight-consolidation approaches (SCoL and Batch TTT) systematically outperform in-context methods, highlighting the importance of updating model parameters rather than relying purely on transient context. Notably, although SCoL is meta-trained on short-context examples, it generalizes effectively to the long-context regime, handling sequences up to twice the training length while remaining competitive with methods that operate on full context. This suggests that our consolidation mechanism captures length-agnostic structure and can extrapolate beyond its training distribution. We hypothesize that further gains for SCoL could be achieved by scaling meta-training to longer sequences or dynamically adapting the strength of the forgetting term $\lambda$ as a function of context length. We refer to \cref{app:lcs-add-res} for additional results, discussions, and standard errors.

\section{Conclusion, limitations, and future work}
\label{sec:conclusions-limitations}

We introduced SCoL, a post-training framework that treats long-context inference as continual context consolidation. 
Instead of keeping new information only in the prompt or an external memory, SCoL post-trains the LLM to generate textual descriptions of where the current context should be written into its own weights.
We formulate consolidation as meta-reinforcement learning problem, where each committed update changes the model state that will generate future update selections.
Across SQuAD knowledge incorporation and LongBench v2 long-context consolidation, SCoL improves acquisition and retention over in-context, summarization, batch, and sequential test-time training baselines.
Analysis of learned selection patterns shows that SCoL encourages sparse update locations that aligns with layers assigned high Fisher information, suggesting that the model learns to route plasticity toward loss-sensitive regions while limiting interference.
Results on LongBench v2  further suggest that selection behavior learned on shorter streams can transfer to longer streaming.
\paragraph{Limitations.} SCoL introduces computational overhead because rewards must be evaluated through inner adaptation. At each context, we sample candidate layer selections, adapt a copy of the current model for each candidate, evaluate acquisition and retention, and then commit the best update. Since each committed update changes the model state used for later contexts, this process is proceed sequentially along the stream, limiting parallelism across contexts.
Additionally, SCoL could exhibit mode collapse over longer IPO updates, concentrating on repeated layer selections and reducing context-dependent diversity (see Appendix~\ref{app:extended_round_dynamics}).
Future work should study policy update algorithms with improved diversity regularization, such as KL control or batch-level entropy penalties.


\clearpage

\bibliography{neurips_2026}
\bibliographystyle{unsrt}
\clearpage
\appendix
\section{Continual knowledge injection: implementation details}
\label{sec:app-cki}
 
This appendix collects the reproducibility details for the
experiments of \Cref{sec:exp-cki}.
 
\subsection{Selection prompt}
\label{sec:app-selection-prompt}
 
The selection prompt is a Qwen chat-template instance shown in
\Cref{fig:selection-prompt}. The prompt is parameterised by the
passage title, the passage context, the implication string, the
per-round budget $B$, and the maximum layer index. We instantiate
$B = 10$ and the maximum layer index at 27 for Qwen2.5-7B.
 
\begin{figure}[h]
\begin{small}
\begin{verbatim}
<|im_start|>system
You are an assistant that selects which Transformer layers to update
so the model best memorizes the given passage and its implications.
<|im_end|>
<|im_start|>user
{title}
{context}
 
Implications:
{implications}
 
Select up to {budget} layer indices most critical for encoding this
knowledge. Each layer is an integer in [0, {max_layer}]. Be selective.
Output ONLY comma-separated integers, no descriptions.<|im_end|>
<|im_start|>assistant
\end{verbatim}
\end{small}
\caption{Selection prompt template.}
\label{fig:selection-prompt}
\end{figure}
 
The model's response is parsed deterministically: all integers in the
output are extracted by regex, filtered to the range $[0, L-1]$,
deduplicated preserving first-seen order, and truncated to the budget
$B$. Each retained layer index is expanded to its seven projection
matrices for adapter attachment. Malformed outputs that yield an
empty parse produce an empty action, which in turn produces a
near-zero reward under the inner adaptation of
\Cref{eq:inner-adapt}; such rollouts are retained in the preference
buffer but are, by construction, almost always the rejected element
of any pair that includes them.
 
\subsection{Implication generation}
\label{sec:app-impl-prompt}
 
Implications are generated by the SEAL-pretrained Qwen checkpoint
after its second self-training round, held frozen throughout all our
experiments. The generation prompt is shown in
\Cref{fig:impl-prompt}.
 
\begin{figure}[h]
\begin{small}
\begin{verbatim}
<|im_start|>system
You are an assistant tasked with analyzing the provided passage and
producing a list of implications derived directly or indirectly from
the content.
<|im_end|>
<|im_start|>user
{title}
{context}<|im_end|>
<|im_start|>assistant
\end{verbatim}
\end{small}
\caption{Implication-generation prompt template, used unmodified
from SEAL.}
\label{fig:impl-prompt}
\end{figure}
 
Sampling uses temperature $0.7$, top-$p$ $0.95$, a maximum of $512$
new tokens, and a single candidate per passage. The resulting
implication string is cached per split and reused verbatim across
all runs (ours and baselines), so implication-generation stochasticity
does not contribute to variance between methods.
 
For inner adaptation, each cached implication string is split on
\texttt{---} and newlines, capped at 30 sequences, and each sequence
is templated as \verb|{title}\n{sequence}|. The raw passage context
is appended as an additional sequence, so the final training set for
one rollout contains at most 31 sequences. This templating is
inherited from SEAL.
 
\subsection{Inner LoRA configuration}
\label{sec:app-inner-lora}
 
\Cref{tab:inner-lora} lists the inner adaptation hyperparameters.
 
\begin{table}[h]
\centering
\small
\caption{Inner LoRA configuration for one candidate rollout.}
\label{tab:inner-lora}
\begin{tabular}{lll}
\toprule
Field & Value  \\
\midrule
Rank $r$                 & 32                 \\
Alpha $\alpha$           & 64                 \\
Dropout                  & 0.0                 \\
Optimiser                & AdamW \cite{loshchilov2018decoupled}  \\
Learning rate            & $2 \times 10^{-4}$  \\
Epochs                   & 10                  \\
Batch size               & 1                   \\
Gradient accumulation    & 1                   \\
Warmup / scheduler       & none                \\
\bottomrule
\end{tabular}
\end{table}
 
The commit policy merges the winning candidate's
adapter into the base weights at the end of each context and
discards the others, so the running base model $\theta_t$ at step
$t$ is a single fully-consolidated Qwen2.5-7B checkpoint rather than
a stack of adapters.
 
\subsection{Outer IPO configuration}
\label{sec:app-outer-ipo}
 
\Cref{tab:outer-ipo} lists the outer preference-optimisation
configuration.
 
\begin{table}[h]
\centering
\small
\caption{Outer-loop IPO configuration.}
\label{tab:outer-ipo}
\begin{tabular}{ll}
\toprule
Field & Value \\
\midrule
Loss                     & IPO \citep{azar2024ipo} \\
$\beta$                  & $0.5$ \\
Preference-margin threshold & $0.05$ (pairs with $|r_w - r_l| < 0.05$ dropped) \\
Pairs per context        & all $\binom{K}{2} = 45$ surviving pairs \\
Optimiser                & AdamW \cite{loshchilov2018decoupled} \\
Learning rate            & \(5\times10^{-6}\)
\\
Gradient accumulation    & 4 \\
Epochs per round         & 2 \\
Reference-policy snap    & at start of each round; re-anchored after update \\
\bottomrule
\end{tabular}
\end{table}
 
The reference policy $\pi_{\mathrm{ref}}$ for the round-$r$ IPO
update is the round-start base model $\pi_{\theta^{(r)}}$ before any
rollout-induced drift, matching the formulation in
\Cref{subsec:meta}. After the IPO step, the updated base model
becomes $\pi_{\theta^{(r+1)}}$ and is used both as the new rollout
substrate for round $r+1$ and as the new reference policy for that
round's outer update.
 
 
\subsection{Reward and judge}
\label{sec:app-reward-judge}
 
The QA judge is a Qwen2.5-7B-Instruct model served as a separate
vLLM instance alongside the editable model. Each judgement receives
a single (question, gold answer, student answer) triple through the
prompt in \Cref{fig:judge-prompt} and returns a binary
decision.
 
\begin{figure}[h]
\begin{small}
\begin{verbatim}
You are a grading assistant. Your job is to determine whether a
student's answer correctly answers the question based solely on the
provided gold answer. Do not use any outside knowledge. The student
answer can include additional information, but it must at least
fully convey the gold answer and must not contradict it. Ignore
style, phrasing, or extra details that do not affect correctness.
Respond ONLY with 'yes' or 'no'.
Question: {question}
Gold answer: {gold}
Student answer: {pred}
Is the student answer correct based solely on the gold answer?
Respond 'yes' or 'no'.
\end{verbatim}
\end{small}
\caption{Judge prompt.}
\label{fig:judge-prompt}
\end{figure}
 
The judge decodes greedily with a maximum of 8 new tokens. A regex
of the form \texttt{\textbackslash b(yes$|$no)\textbackslash b} is
run over the output and the last match is taken as the decision.
Outputs whose last match is not ``yes'' (including outputs that
produce no match at all) are scored zero. Accuracy on a single
$\mathcal{Q}_t$ is the mean of the binary decisions over the
roughly five questions in that passage.
 
 
\subsection{Metric formulas}
\label{sec:app-metrics}
 
All main-text metrics are summary statistics of the sequential-edit
accuracy matrix $M \in [0,1]^{N \times N}$, with $N = 100$ the
length of the evaluation stream. Writing $M_{i,j}$ for the accuracy
on $\mathcal{Q}_j$ after editing through $c_i$ ($j \le i$), we
report
\begin{align*}
\text{Immediate acquisition} \;&=\; \tfrac{1}{N}\textstyle\sum_{i} M_{i,i}, \\
\text{Retention} \;&=\; \tfrac{1}{N-1}\textstyle\sum_{j < N-1} M_{N-1, j}, \\
\end{align*}
 
\subsection{Stream and round configuration}
\label{sec:app-stream}
 
\Cref{tab:stream-config} lists the continual-stream configuration.
 
\begin{table}[h]
\centering
\small
\caption{Stream configuration for \Cref{sec:exp-cki}.}
\label{tab:stream-config}
\begin{tabular}{ll}
\toprule
Field & Value \\
\midrule
Training passages per round $T$ & 50 \\
Rounds reported in main text $R$ & 2 (plus round 0) \\
Rollouts per context $K$         & 10 \\
Selector sampling temperature    & 1.0 \\
Shuffle seed                     & 42 (applied once before slicing) \\
Round-to-passage mapping         & R1: passages $[0,50)$; R2: passages $[50,100)$ \\
Evaluation stream length         & 100 passages (fixed order) \\
Past-query subsampling           & none \\
\bottomrule
\end{tabular}
\end{table}
 
The training cache contains 250 shuffled passages; \Cref{sec:exp-cki}
uses the first two disjoint slices of 50 passages each. The remaining
150 passages are reserved for extended-rounds diagnostics
(\Cref{sec:app-granularity}) and are not used by any main-text
result.
 
\subsection{Baseline implementation details}
\label{sec:app-baselines}
 
\paragraph{{QA Prompting}.}
Prompting the base model with question only.
 
\paragraph{{Batch LoRA FT}.}
LoRA attached to all 28 layers, trained jointly on the concatenated
implications of all 100 evaluation passages for 10 epochs at learning
rate $2 \times 10^{-4}$.
 
\paragraph{{SEAL}.}
The SEAL knowledge-incorporation method applied to our continual
stream, implemented as a sequential LoRA update per passage with the inner
LoRA configuration of \Cref{tab:inner-lora}.
 
\paragraph{SCoL, both variants.}
Identical configurations; the only difference is the value of the
reward weight $\lambda$.

\subsection{Additional results}
\label{app:addi-results-cki}
See standard errors in \Cref{tab:seed_sweep}
\begin{table}[h]
\centering
\caption{Standard errors for round-2 SQuAD sequential editing results over $K=3$ inner-training seeds.}
\label{tab:seed_sweep}
\begin{tabular}{lcc}
\toprule
Method & Immediate acc.\ & Retention  \\
\midrule
SEAL & $0.017$ & $0.012$ \\
SCoL$_{\lambda=0}$                      & $0.020$ & $0.016$  \\
SCoL                   & $0.008$ & $0.013$  \\
\bottomrule
\end{tabular}
\end{table}

\begin{figure}[h]
    \centering
    \includegraphics[width=1\linewidth]{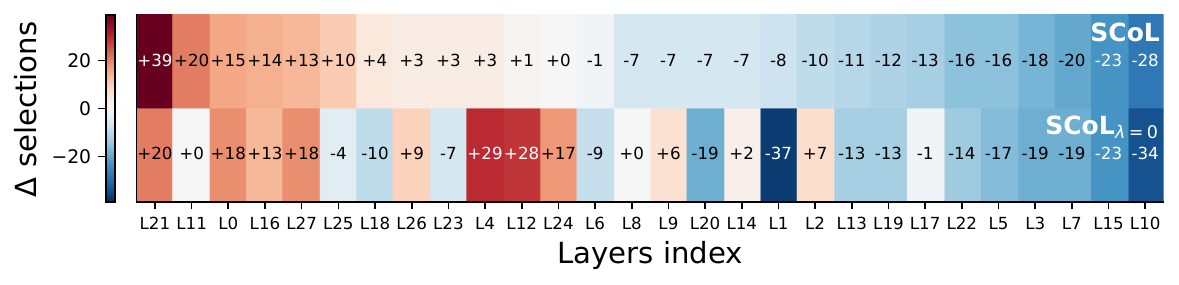}
    \caption{Per-layer change in selection count between the initial and final IPO updates. Layers are sorted by SCoL's largest gain (left) to largest drop (right). SCoL (top, full reward) concentrates onto a handful of layers and abandons most others, while SCoL$_{\lambda=0}$ (bottom, acquisition only) shifts selection mass differently.}

    \label{fig:selection_delta}
\end{figure}

\subsection{Compute resources}
\label{sec:app-compute}
 
All experiments ran on NVIDIA RTX PRO 6000 Blackwell GPUs (96\,GB each).
A single training round ($K{=}10$ rollouts, $T{=}50$ passages)
took approximately $4.3$ hours per configuration. A single sequential
matrix evaluation on the $N{=}100$ validation stream took $50$ to $110$
minutes on one GPU.

\subsection{Outer Loop Algorithm Sweep: IPO, DPO, and ReST}
\label{app:outer_algo_sweep}

We sweep three outer loop policy update algorithms while holding the inner loop, reward, stream, and rollout budget fixed. Each selector is trained on one round of $T{=}50$ passages with $K{=}10$ rollouts per passage, layer granularity, $\lambda{=}1$, and evaluated on the same $N{=}100$ validation stream. We compare ReST \citep{gulcehre2023rest}, DPO \citep{rafailov2023dpo}, and IPO from Eq.~(8) \citep{azar2024ipo}. For each algorithm, we sweep learning rate, gradient accumulation, and the algorithm specific parameters. All other hyperparameters match Table~2.

Following the mode concentration concern in Section~3.2, we report selector uniqueness as a diagnostic. For each evaluated selector, we count the number of \emph{distinct} selection completions emitted across the 100 validation passages, denoted uniq/100. We also report top1\%, the share of the most common completion. uniq/100 is necessary but not sufficient for a useful selector, since high uniqueness can also reflect noisy selections. We therefore read it jointly with immediate accuracy.

\begin{table}[t]
\centering
\small
\setlength{\tabcolsep}{4pt}
\renewcommand{\arraystretch}{1.08}
\caption{Outer loop algorithm sweep. Retention uses the JSON summary key from \texttt{seq\_eval.json}, not the paper retention metric in Section~A.6.}
\label{tab:outer_algo_sweep}
\begin{tabular}{llcccc}
\toprule
\textbf{Algorithm} & \textbf{Key hyperparameters} & \textbf{Imm. Acc.} & \textbf{Retention} & \textbf{uniq/100} & \textbf{top1\%} \\
\midrule
IPO  & $\beta{=}0.5$                         & 36.60\% & 69.30\% & 65 & 8  \\
IPO  & $\beta{=}0.5$, ep$=2$, ga$=4$, lr$=5{\times}10^{-6}$ & 38.00\% & 65.80\% & 73 & 5  \\
IPO  & $\beta{=}0.1$                         & 33.90\% & 54.30\% & 81 & 3  \\
IPO  & $\beta{=}0.01$                        & 34.20\% & 74.40\% & 79 & 7  \\
DPO  & $\beta{=}0.1$                         & 35.90\% & 42.20\% & 83 & 8  \\
DPO  & $\beta{=}0.5$                         & 33.70\% & 66.30\% & 77 & 6  \\
ReST & lr$=5{\times}10^{-5}$, ga$=2$, $k{=}1$ & 35.90\% & 74.20\% & 59 & 9  \\
ReST & lr$=3{\times}10^{-4}$, ga$=10$, $k{=}1$ & 41.00\% & 27.30\% & 67 & 7  \\
ReST & lr$=3{\times}10^{-4}$, ga$=2$, $k{=}3$ & 12.90\% & 57.80\% & 14 & 61 \\
\bottomrule
\end{tabular}
\end{table}

IPO is the only update rule that jointly preserves uniq/100 above roughly 65 and reaches the strongest immediate accuracy, with 36.60\% to 38.00\% immediate accuracy and 65.80\% to 69.30\% retention at $\beta{=}0.5$. ReST shows the failure mode discussed in Section~3.2. Its aggressive setting reaches 41.00\% immediate accuracy, but retention drops to 27.30\%, while the most collapsed ReST setting has only 14 distinct completions and top1\% of 61\%. This trend is consistent with mode concentration under cross entropy style updates \citep{omahony2024modecollapse,li2025preservediversity}. DPO retains high uniq/100, often between 77 and 83, but underperforms IPO on the joint of immediate accuracy and retention. This sweep motivates IPO with $\beta{=}0.5$ as the headline outer loop algorithm, and we use this configuration for all main text results.

\subsection{Granularity Comparison: Layer, Module, and Per Projection}
\label{app:granularity_comparison}

The action space in Eq.~(4) ranges over $L$ structural units of the base model. We instantiate three choices in the same codebase, with budgets chosen to keep the per rollout adapter parameter count comparable. The layer setting uses $L{=}28$ and budget $B{=}10$, where each selected layer attaches LoRA adapters to seven projection modules. The module setting uses $L{=}56$ and budget $B{=}20$, where each layer is split into attention and MLP units. The per projection setting uses $L{=}196$ with budgets $B{=}70$ and $B{=}100$, where each q, k, v, o, gate, up, and down projection is a separate slot. All settings share the inner LoRA configuration in Table~1, the IPO configuration in Table~2, the same stream, $K{=}10$, and both reward variants.

\begin{table}[t]
\centering
\small
\setlength{\tabcolsep}{4pt}
\renewcommand{\arraystretch}{1.08}
\caption{Round 2 granularity comparison. SCoL uses the full reward. SCoL$_{\lambda=0}$ removes the forgetting term from the reward.}
\label{tab:granularity_r2}
\begin{tabular}{lccccccc}
\toprule
\multirow{2}{*}{\textbf{Granularity}} & \multirow{2}{*}{$\mathbf{B}$} &
\multicolumn{3}{c}{\textbf{SCoL}} &
\multicolumn{3}{c}{\textbf{SCoL$_{\lambda=0}$ }} \\
\cmidrule(lr){3-5}
\cmidrule(lr){6-8}
& & \textbf{Imm.} & \textbf{Ret.} & \textbf{uniq/100} & \textbf{Imm.} & \textbf{Ret.} & \textbf{uniq/100} \\
\midrule
Layer          & 10  & 35.40\% & 20.46\% & 82  & 34.30\% & 14.64\% & 85 \\
Module         & 20  & 35.10\% & 20.21\% & 100 & 36.61\% & 19.45\% & 95 \\
Per projection & 70  & 24.70\% & 16.60\% & 99  & 31.40\% & 14.50\% & 96 \\
Per projection & 100 & 29.80\% & 20.00\% & 98  & 29.50\% & 17.00\% & 95 \\
\bottomrule
\end{tabular}
\end{table}

Layer and module granularity yield comparable Round 2 immediate accuracy, around 35.00\% across both reward variants, but the layer configuration is more stable across rounds and is therefore used in the main text. Per projection granularity underperforms by roughly five to ten absolute points in immediate accuracy, despite its larger nominal action space. Since uniq/100 remains high, often above 90, the failure is better explained as selector \emph{noise} rather than selector collapse.

This pattern suggests that action space expressivity is bottlenecked by structured generation. As granularity becomes finer, the policy must emit longer and more constrained outputs, moving from comma separated layer indices to module pairs and then projection pairs. With budgets of 70 or 100, the policy must produce a long list of valid pairs whose indices and projection names obey the parser grammar. A 7B base policy struggles with this output length and schema fidelity, so the effective entropy of useful rollouts shrinks even as the nominal action space grows.

Two future directions follow. First, the gap between layer and per projection granularity may narrow with stronger base policies, since long structured generation is plausibly scaling sensitive \citep{wei2022emergent}. We state this only as a hypothesis. Second, the outer loop update can be made more diversity preserving at extreme structured lengths, either through entropy regularization \citep{li2025preservediversity} or through an objective that scores list \emph{compositions} rather than independent positions. We leave both directions to future work.

\subsection{Extended Round Dynamics and Mode Collapse}
\label{app:extended_round_dynamics}

We also continue the outer loop beyond the two rounds reported in the main text. This experiment is diagnostic rather than a headline result. We run two additional rounds, R3 and R4, for layer and module granularity under both reward variants. The inner loop, IPO objective, stream, and budget are unchanged. After each round, we record uniq/100 and immediate accuracy.

\begin{table}[h]
\centering
\small
\setlength{\tabcolsep}{4pt}
\renewcommand{\arraystretch}{1.08}
\caption{Extended round diagnostics. Each cell reports immediate accuracy and uniq/100.}
\label{tab:extended_round_dynamics}
\begin{tabular}{lcccc}
\toprule
\textbf{Granularity and reward} & \textbf{R1} & \textbf{R2} & \textbf{R3} & \textbf{R4} \\
\midrule
Layer, SCoL$_{\lambda=0}$
& 34.31\% / 64 & 35.40\% / 82 & 33.50\% / 39 & 36.50\% / 5 \\
Layer, SCoL
& 35.16\% / 85 & 34.30\% / 85 & 37.50\% / 69 & 39.10\% / 33 \\
Module, SCoL$_{\lambda=0}$
& 27.30\% / 93 & 35.10\% / 100 & 6.70\% / 99 & 11.00\% / 63 \\
Module, SCoL
& 31.40\% / 96 & 36.60\% / 95 & 34.30\% / 76 & 27.60\% / 7 \\
\bottomrule
\end{tabular}
\end{table}

uniq/100 declines with additional rounds, and the full reward variant collapses faster than the acquisition only variant. On layer granularity, SCoL  drops from 82 distinct completions at R2 to 5 at R4, while SCoL$_{\lambda=0}$  retains 33 at R4. The module setting shows the same direction, with the full reward variant also collapsing in immediate accuracy to 6.70\% at R3.

Round 2 should therefore be read as an early stopping point for selector quality, not as the asymptotic behavior of the fixed objective. With the same IPO loss and buffer construction, later rounds concentrate probability mass on a small number of high reward actions observed during training and reduce exploration. This is the mode concentration mechanism flagged in Section~3.2 and observed in cross entropy and preference style fine tuning of LLMs \citep{omahony2024modecollapse,li2025preservediversity}. For this reason, the main text reports R0 to R2 and treats further round scaling as a regularization problem for future work.

\subsection{Layerwise Fisher computation}
\label{app:fisher-computation}

For each passage \(c\), let \(\mathcal{I}(c)\) denote its generated implication lines. Fisher scores are computed on the running consolidated base model before attaching a new LoRA adapter for the passage. For each implication line \(x \in \mathcal{I}(c)\), we compute the autoregressive sum loss
\[
\mathcal{L}(x;\theta)=-\sum_{t=1}^{|x|}\log p_{\theta}(x_t\mid x_{<t}),
\]
where \(\theta\) denotes the current consolidated base parameters. We estimate diagonal Fisher importance by squared gradients:
\[
F_j(c)=\frac{1}{|\mathcal{I}(c)|}\sum_{x\in \mathcal{I}(c)}
\left(\frac{\partial \mathcal{L}(x;\theta)}{\partial \theta_j}\right)^2 .
\]
Let \(\mathcal{P}_{\ell}\) be the parameters of the seven base projection matrices in layer \(\ell\) that are eligible for LoRA attachment. The layerwise Fisher score is
\[
F_{\ell}(c)=\sum_{j\in \mathcal{P}_{\ell}}F_j(c),
\qquad \ell\in\{0,\ldots,27\}.
\]
This yields a 28 dimensional Fisher vector \(F(c)\) for each passage. These scores are used only for analysis and are never provided to the LLM when it generates layer selections.
\section{Long-context Consolidation}
\label{app:lcs}

\subsection{Hyperparameters}
\label{app:lcs-hyperparameters}
\subsubsection{Inner LoRA configuration}
We retain the same LoRA configuration from \Cref{app:addi-results-cki} and \Cref{tab:inner-lora}. Additionally, all our chunks are of a fixed $2048$ token length calculated using the \texttt{Qwen2.5-7B} tokenizer and we create a training set by further chunking these into $128$ token size subchunks. 

\subsubsection{Outer IPO configuration}
We retain mostly the same IPO configuration from \Cref{app:addi-results-cki} and \Cref{tab:outer-ipo}. We only change the preference margin threshold from $0.05$ to $0.3$. That is, given candidates $c_1, c_2$ with rewards $r_1, r_2$ respectively and $r_1>r_2$, we put $\{c_1,c_2\}$ in the IPO buffer iff $r_1-r_2 > 0.2$.

\subsection{Implementation Details}
\label{app:lcs-imp-det}

\paragraph{SCoL.}
We retain the majority of the setup from \Cref{sec:app-cki}, with the modifications presented in \Cref{app:lcs-hyperparameters}
\paragraph{In-context settings.}
We use the base \texttt{Qwen2.5-7B} model with temperature $= 1$ implemented using vLLM. 

For \textbf{Summarization}, we prompt \texttt{Kimi-K2.6} using the template shown in \Cref{fig:summarization-prompt}. We set \texttt{maxTokens} to 10{,}000 to discourage degenerate behavior where the model copies large portions of the original passage instead of compressing it.

\begin{figure}[h]
\begin{small}
\begin{verbatim}
Please provide a dense and concise summary of the following passage in under 
{maxTokens} tokens keeping all the important information:\n\n{context}\n\nSummary:
\end{verbatim}
\end{small}
\caption{Summarization prompt used for Kimi-K2.6.}
\label{fig:summarization-prompt}
\end{figure}

\paragraph{Test-time training.}
For both Batch TTT and Sequential FT, we attach LoRA adapters to all 28 transformer layers. Training is performed either jointly on the full passage (Batch TTT) or incrementally over sequential chunks (Sequential FT). Each passage is evaluated using freshly initialized LoRA adapters trained under the same inner-loop test-time training configuration as given in \Cref{app:lcs-hyperparameters}.

\subsubsection{Dataset Processing}
Our goal is to evaluate whether long contexts can be continually consolidated into model parameters. We therefore use LongBench-v2, which contains naturally long documents (e.g., books and multi-section texts) paired with reasoning-based question-answer tasks.

Each document is treated as an independent context and processed as a stream of 2048-token segments, simulating a setting where information arrives sequentially. For training, we use 10 passages with lengths between 16k and 32k tokens. We sample 40 different passages for evaluation and an additional 20 passages of lengths between 32k and 64k tokens.

Tokenization and segmentation are performed using the \texttt{Qwen2.5-7B} tokenizer. All dataset splits are created with a fixed random seed of 42 for reproducibility.

\subsubsection{Compute Resources}
All experiments for this section are conducted on NVIDIA H200 GPUs with 8 CPUs. Meta-training (2 rounds over 5 passages each) takes approximately 8 hours. Evaluation over the full benchmark (40 short and 20 long passages) requires approximately 6 hours.

\subsection{Additional Results}
\label{app:lcs-add-res}

\paragraph{Passage Log-likelihoods.}
We further compare the final Log-Likelihoods between batch TTT, sequential FT, and SCoL in \Cref{tab:lcs-logprobs-app}.
Consistent with the main results, Batch TTT achieves the highest likelihood in both regimes ($-1.26$), indicating that full-sequence optimization yields the best fit to the data distribution. In contrast, Sequential FT performs substantially worse ($-2.32$ short, $-2.51$ long), reinforcing that naïve online updates lead to degraded representations and poor predictive calibration.
\begin{wraptable}[10]{r}{0.35\textwidth}
\centering
\small
\caption{Long-context consolidation comparison of final log-likelihoods across short and long regimes.}
\begin{tabular}{lcc}
\toprule
\textbf{Method} & \textbf{short ($\uparrow$)} & \textbf{long ($\uparrow$)} \\
\hline
\noalign{\vskip 1pt}
Batch TTT           & \textbf{-1.26} & \textbf{-1.26} \\
Sequential FT       & -2.32 & -2.51 \\
\hline
\noalign{\vskip 1pt}
\textbf{SCoL}     & -1.44 & -1.82 \\
\bottomrule
\end{tabular}
\label{tab:lcs-logprobs-app}
\end{wraptable}

SCoL consistently improves over Sequential FT in both regimes ($-1.44$ vs. $-2.32$ in short, $-1.82$ vs. $-2.51$ in long), indicating that structured consolidation mitigates the degradation in likelihood associated with continual updates. Notably, the gap between SCoL and Batch TTT is smaller in the short-context regime (0.18) than in the long-context regime (0.56), mirroring the trends observed in accuracy.

We hypothesize that this widening gap reflects the same stability–plasticity tradeoff observed earlier. In the short-context setting, SCoL’s forgetting-aware objective enables updates that remain close to the optimal batch solution, yielding likelihoods comparable to Batch TTT. However, as context length increases, the constraint imposed by the forgetting term may limit the model’s ability to fully adapt to the extended sequence, leading to lower likelihood relative to Batch TTT.

Overall, these results support the conclusion that SCoL provides a strong approximation to batch retraining while maintaining stability, substantially improving over naïve sequential updates, though some loss in optimality remains when full-context adaptation is required.

\begin{figure}
    \centering
    \includegraphics[width=0.7\linewidth]{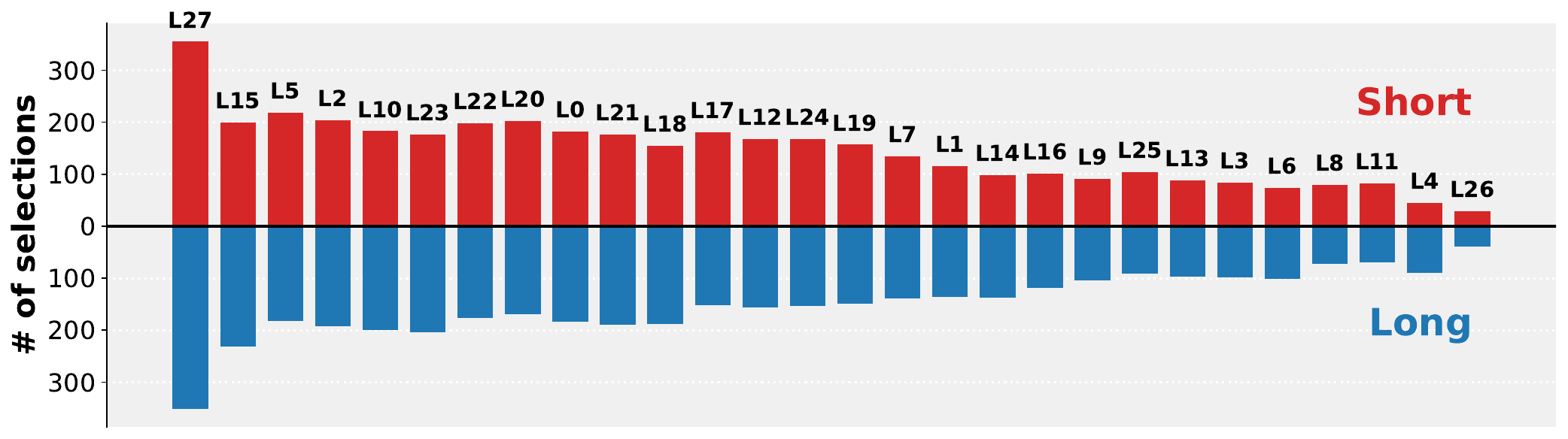}
    \caption{Layer selections for SCoL for the long context setting for short and long regimes.}
    \label{fig:lcs_layer_selections}
\end{figure}

\paragraph{Layer selection frequencies.}
We also show the top layer selection frequencies for our Long Context Consolidation Setting in \Cref{fig:lcs_layer_selections}. Similar to the results in \Cref{sec:exp-cki}, we find that L27 is the selected most often but here we find that L26 and L25 are selected relatively less. Additionally, the middle layers are selected least often matching the SQuAD setting.

\paragraph{Full results table with standard errors.} 
\begin{table}[th]
\centering
\small
\caption{Long-context consolidation comparison across two context-length regimes (short: 16k--32k and long: 32k--64k tokens).}
\begin{tabular}{lcc}
\toprule
\textbf{Method} & \textbf{short} & \textbf{long} \\
\hline
\noalign{\vskip 1pt}
Base        & 32.5$_{\pm 0.4}$\% & 29.5$_{\pm 0.5}$\% \\
Full Context    & 38.0$_{\pm 0.5}$\% & 35.5$_{\pm 0.8}$\% \\
Summarization   & 39.3$_{\pm 0.5}$\% & 36.0$_{\pm 0.7}$\% \\
\hline
\noalign{\vskip 1pt}
Batch TTT           & \underline{39.5$_{\pm 0.5}$\%} & \textbf{41.5$_{\pm 0.5}$\%} \\
Sequential FT       & 36.5$_{\pm 0.8}$\% & 23.0$_{\pm 1.0}$\% \\
\hline
\noalign{\vskip 1pt}
\textbf{SCoL}     & \textbf{42.3$_{\pm 0.8}$\%} & \underline{37.0$_{\pm 1.0}$\%} \\
\bottomrule
\end{tabular}
\label{tab:lcs-results-variance-app}
\end{table}

We provide the full results table for the long context experiments, along with the standard errors in table \Cref{tab:lcs-results-variance-app}.
\section{Broader Impacts}
\label{sec:broader_impacts}
The proposed SCoL framework enables long-context internalization through persistent weight updates, offering significant improvements in computational efficiency and long-term agent memory. However, this shift from transient context to persistent weights introduces privacy risks, as consolidated sensitive information may be harder to redact. Furthermore, continual self-adaptation may inadvertently reinforce biases or internalize misinformation present in the input stream.

\section{Licenses}

\label{sec:licenses}
We provide details regarding the licenses of external assets used in this work, presented in Table \ref{tab:licenses}.

\begin{table}[h]
\centering
\small
\setlength{\tabcolsep}{3pt}
\renewcommand{\arraystretch}{1.2}
\captionsetup{skip=8pt}
\begin{tabular}{p{0.25\textwidth} p{0.2\textwidth} p{0.45\textwidth}}
\toprule
\textbf{Asset} & \textbf{URL} & \textbf{License} \\
\midrule
\multicolumn{3}{l}{\textbf{Datasets}} \\
\midrule
SQuAD \cite{rajpurkar2016squad} & \href{https://rajpurkar.github.io/SQuAD-explorer/}{Link} & Creative Commons Attribution-ShareAlike 4.0 International \\
LongBench-v2 \cite{bai2025longbenchv2} & \href{https://huggingface.co/datasets/zai-org/LongBench-v2}{Link} & MIT License \\
\midrule
\multicolumn{3}{l}{\textbf{Models}} \\
\midrule
Qwen Models \cite{qwen} & \href{https://github.com/QwenLM/Qwen}{Link} & Apache License 2.0\textsuperscript{*} \\
\bottomrule
\end{tabular}
\caption{Licenses for assets used in our experiments. \textsuperscript{*}The majority of the Qwen series (e.g., Qwen1.5, Qwen2, Qwen2.5) are licensed under Apache 2.0, with a few exceptions for specific legacy sizes governed by the Tongyi Qianwen License Agreement.}
\label{tab:licenses}
\end{table}





\end{document}